\PassOptionsToPackage{table}{xcolor}
\documentclass[acmsmall,nonacm]{acmart}
\acmJournal{TELO}
\graphicspath{{figures/}}

\begin{document}
\title[On the Non-Linearity of Symbolic Regression Problems and How to Exploit It with Model-based Semantic GP]{TSGP-2: On the Non-Linearity of Symbolic Regression Problems and How to Exploit It with Model-based Semantic Genetic Programming}

\author{Philipp Anthes}
\email{anthes@uni-mainz.de}
\orcid{0009-0009-8383-6551}
\affiliation{%
  \institution{Johannes Gutenberg University}
  \city{Mainz}
  \country{Germany}
}
\author{Alina Geiger}
\orcid{0009-0002-3413-283X}
\affiliation{%
  \institution{Johannes Gutenberg University}
  \city{Mainz}
  \country{Germany}}
\author{Martin Briesch}
\orcid{0000-0002-8209-1465}
\affiliation{%
  \institution{Johannes Gutenberg University}
  \city{Mainz}
  \country{Germany}}
\author{Franz Rothlauf}
\orcid{0000-0003-3376-427X}
\affiliation{%
  \institution{Johannes Gutenberg University}
  \city{Mainz}
  \country{Germany}}
\renewcommand{\shortauthors}{Anthes et al.}

\begin{abstract}
Model-based variation operators such as the one used in Transformer Semantic Genetic Programming (TSGP) are affected by possible biases of their training data. TSGP is trained on programs generated by Genetic Programming (GP) runs on linear target functions, which leads to a bias toward linear offspring. This bias can hinder search performance on problems that require non-linear solutions. This work has two goals: first, we investigate the linearity of standard benchmark problems including the Feynman and real-world problems. While most of the structure of real-world problems can be well approximated by linear functions, the Feynman problems exhibit varying levels of non-linear structure. Thus, a proper variation operator of a search method should be able to create both types of programs, linear as well as non-linear ones. Second, we propose TSGP-2, which follows the functionality of TSGP but uses as training data programs from GP runs on both linear and non-linear target functions. As a result, the training data includes a lower share of programs that behave linearly. With a higher share of non-linear structures, TSGP-2 outperforms TSGP on problems with non-linear structure, while performing similarly on linear problems. In addition, TSGP-2 is less biased toward linear offspring and is able to propose a higher number of meaningful non-linear offspring earlier during search, which coincides with better solutions. The findings underscore the importance of training data for a learned variation operator and emphasize that the biases of search operators should be aligned with the problems they are applied to.
\end{abstract}



\maketitle

\section{Introduction}
\label{sec:Introduction}
The goal of symbolic regression (SR) is to find a mathematical function whose behavior, its \emph{semantics}, fits a given target behavior as closely as possible. In genetic programming (GP) these functions are evolved in an evolutionary search process through selection and variation. In particular, the variation operators have a major influence on which behavior is found and how quickly the search converges to a good solution~\cite{vanneschi2014semantic,moraglio2012gsgp,uy2011semantic}.

Standard operators such as subtree mutation and subtree crossover~\cite{koza1992gp,poli2008fieldguide} randomly vary parts of the function. They thereby ignore how a variation affects a function's semantics~\cite{moraglio2012gsgp}. Consequently, variation often produces offspring that are either semantically identical to their parents or substantially different from them, both of which harm search efficiency~\cite{mcphee2008semantic,beadle2009sdm,uy2009semantics}. Semantic GP addresses this with variation operators that control the semantic similarity between offspring and parents~\cite{vanneschi2014semantic}. Some approaches such as Geometric Semantic GP (GSGP)~\cite{moraglio2012gsgp} and SLIM\_GSGP~\cite{vanneschi2024slim} work with pre-defined syntactic rules that define how a parent should be varied to obtain an offspring that is semantically similar. 

Instead of relying on fixed rules, the recently proposed Transformer Semantic GP (TSGP)~\cite{anthes2025transformer} uses a trained machine learning model as a semantic variation operator. This model is trained using pairs of semantically similar functions to learn how a syntactic change affects the semantics of a function. TSGP is able to produce meaningful offspring, effectively guiding the search process. As a result, it finds better solutions for a wide range of SR problems compared to standard GP, SLIM\_GSGP, GSGP, and other model-based approaches~\cite{anthes2026dimensional,anthes2026syntactic}. An open and unsolved question is why the performance improvements of TSGP are not consistent across all problems~\cite{anthes2026dimensional}. 

Recent work on pretrained SR models has shown that learned models are strongly biased toward their training distribution~\cite{sato2026reproduction,voigt2025generalization}. In TSGP, the model is trained on programs generated during GP runs that search for linear target functions~\cite{anthes2026dimensional}. It is still unclear whether and how the choice of training data affects the offspring proposed by the model, and, ultimately, search performance. Understanding these effects is crucial if the model is to be applied to a wide range of problems with different characteristics. While a linear function might already explain most of the target behavior for some problems, others have shapes that a linear function cannot sufficiently approximate. Instead, non-linear functions are required to solve these problems~\cite{udrescu2020feynman}. If the variation operator of TSGP is indeed biased toward linear offspring, this would harm search performance on non-linear problems.

Therefore, we analyze in this work the problem structure of synthetic as well as real-world benchmark problems regarding the linearity of their behavior. Motivated by these findings, we generate training data based on both linear and non-linear target functions. With this training data, we train a novel model-based variation operator which we call TSGP-2. We compare the search performance of TSGP-2 with TSGP and standard GP across a wide range of synthetic and real-world problems and trace the differences back to the search behavior and the variation operators.

We find that while synthetic Feynman problems exhibit clear non-linear structures, most of the analyzed real-world problems can largely be explained with linear approximations. Only four of the $13$ real-world problems exhibit clear non-linear structures. Furthermore, we confirm that the training data of TSGP consists mainly of programs that behave linearly ($83\%$), compared to TSGP-2 ($37\%$).

With this less linear training data, TSGP-2 achieves the best mean rank on both the Feynman problems ($1.19$, compared to $1.88$ for TSGP and $2.94$ for standard GP) and the real-world problems ($1.46$, compared to $1.92$ and $2.62$). On the Feynman problems, it significantly outperforms TSGP on $7$ of the $16$ problems, while TSGP is never significantly better than TSGP-2. This gap in performance between TSGP and TSGP-2 is evident on problems with clear non-linear structures, whereas on problems without them, both reach comparable solution quality.

Our analysis traces this advantage back to the training data. On the Feynman problems, the best programs of TSGP-2 leave the linear region earlier than those of TSGP (generation $11$ vs.\ $16$), after which its error advantage over TSGP grows. During the search, TSGP-2 proposes non-linear offspring earlier and more often, and selection keeps them, whereas TSGP keeps proposing linear offspring that selection discards. Standard GP proposes the most strongly non-linear offspring, but selection rarely keeps them due to their low quality. Thus, the learned TSGP and TSGP-2 operators propose non-linear programs that are more relevant for search. 

Section~\ref{sec:Related_Work} describes related work on semantic GP and model-based approaches and discusses training-data bias in learned models. Section~\ref{sec:MeasuringLinearity} defines our linearity measure $\lambda$ and analyzes the linearity of the benchmark problems. Section~\ref{sec:TSGP} describes the TSGP approach, introduces TSGP-2, and compares the linearity of their training data. Section~\ref{sec:Search} compares the search performance of TSGP-2, TSGP, and standard GP, and Section~\ref{sec:Analysis} traces the differences back from the best programs and the offspring to the operators. Section~\ref{sec:Conclusion} concludes this paper.

\section{Related Work} 
\label{sec:Related_Work}
\subsection{Semantic GP}
Standard variation operators apply syntactic changes to an expression with unpredictable semantic effects~\cite{vanneschi2014semantic}. Therefore, semantic variation operators have been proposed that control the semantic effect of a variation to better guide the search process. We focus on semantic mutation operators, which can be divided into indirect and direct approaches~\cite{vanneschi2014semantic}. 

Indirect approaches keep the standard mutation and filter its offspring by their semantics. Semantically driven mutation~\cite{beadle2009sdm} rejects offspring that are semantically identical to their parent, and semantic similarity-based mutation~\cite{uy2009semantics} accepts only offspring whose semantic distance to the parent lies within a given range. In contrast, direct approaches construct offspring with a controlled semantic effect. GSGP~\cite{moraglio2012gsgp} follows a pre-defined rule by adding a random substructure to the parent. The output of this substructure is scaled by a mutation step, so that the semantic change is bounded. As a result, the offspring lies within a controlled semantic distance to its parent. However, because every offspring contains its entire parent, expressions grow quickly. SLIM\_GSGP~\cite{vanneschi2024slim} adds a deflate mutation that removes a previously added substructure so that individuals can also shrink. Subsequent work explores further variants~\cite{vanneschi2025exploring}, the choice of the mutation step~\cite{farinati2025mutationstep}, and the extension of SLIM\_GSGP with crossover~\cite{pietropolli2025introducing}.

TSGP~\cite{anthes2025transformer,anthes2026dimensional} is also a direct approach, but replaces pre-defined rules with a trained machine learning model that proposes, for a given parent, an offspring of similar semantics.

\subsection{Model-based GP}
Model-based GP replaces standard variation operators with a model learned from data with the goal of passing on promising substructures to the offspring rather than disrupting them through random variation~\cite{kim2014pmbgp,virgolin2021gomea}. These include estimation-of-distribution algorithms (EDAs), which estimate a probability distribution over the selected individuals in every generation and sample the next population from it~\cite{larranaga2002eda,hauschild2011introduction}. They differ mainly in the dependencies their model can express. 
PIPE~\cite{salustowicz1997pipe} gives each node position of a prototype tree an independent distribution over the instruction set, stochastic grammar-based GP~\cite{ratle2001avoiding} learns probabilities for the production rules of a grammar, and N-gram GP~\cite{poli2008ngram} conditions each instruction of a linear program on the two preceding ones. DAE-GP~\cite{wittenberg2020daegp} uses an LSTM denoising autoencoder that generates offspring by corrupting a parent and reconstructing it, with the corruption strength controlling the trade-off between exploration and exploitation~\cite{wittenberg2022syntax,wittenberg2023daegp}. 
In contrast, GP-GOMEA~\cite{virgolin2017scalable,virgolin2021gomea} does not sample from its model. Instead, it learns which node positions should be varied together and uses this linkage model to mix these groups of nodes between solutions.

TSGP differs from these approaches in two ways. First, its model is fitted once on separate training data before any search, rather than on the population of the current run, and second, the model learns how to generate semantically similar offspring.

\subsection{Bias in Learned Models} 
In symbolic regression, pretrained models~\cite{biggio2021nesymres,kamienny2022e2e,vastl2024symformer} are trained on millions of expressions sampled from a random expression generator~\cite{lample2019deep}, each paired with its evaluation on data. Given a new dataset, they predict an expression directly by decoding from the model instead of using an evolutionary search and either propose it as a skeleton function whose constants are fitted afterwards~\cite{biggio2021nesymres} or include the constants directly~\cite{kamienny2022e2e}. This makes them fast and accurate on problems that resemble their training data, but inaccurate for problems that are not contained in the training set. Sato and Sato~\cite{sato2026reproduction} call this reproduction bias and measure it as the share of generated expressions whose structure already occurs in the training data. Voigt et al.~\cite{voigt2025generalization,voigt2026beyond} report that such models hold their accuracy inside the pre-training distribution and degrade outside it. In other areas of supervised learning, such as computer vision and natural language processing, it is also well studied that a model inherits the biases of its training data~\cite{geirhos2019texture,torralba2011bias, mccoy2019hans,gururangan2018annotation}. 

Like these models, TSGP is trained via supervised learning, but it operates as a variation operator that conditions each proposed offspring on a parent. Consequently, any bias inherited from its training data influences the direction in which an offspring is moved away from its parent within the semantic space.

\section{Linearity of Solutions}
\label{sec:MeasuringLinearity}
This section suggests a measure for linearity and applies it to programs and standard test problems. Section~\ref{sec:Lambda} introduces $\lambda$, which measures the linearity by how much of its behavior a linear function explains. We use $\lambda$ throughout this work to measure the linearity of the training data, of the offspring, and of the best programs of a search. Section~\ref{sec:Benchmarks} applies the same idea to the targets of common SR benchmark problems, which shows whether solving them requires linear or non-linear solutions.

\subsection{A Measure of Linearity}
\label{sec:Lambda}
\begin{figure}[t]
\centering
\includegraphics[width=\linewidth]{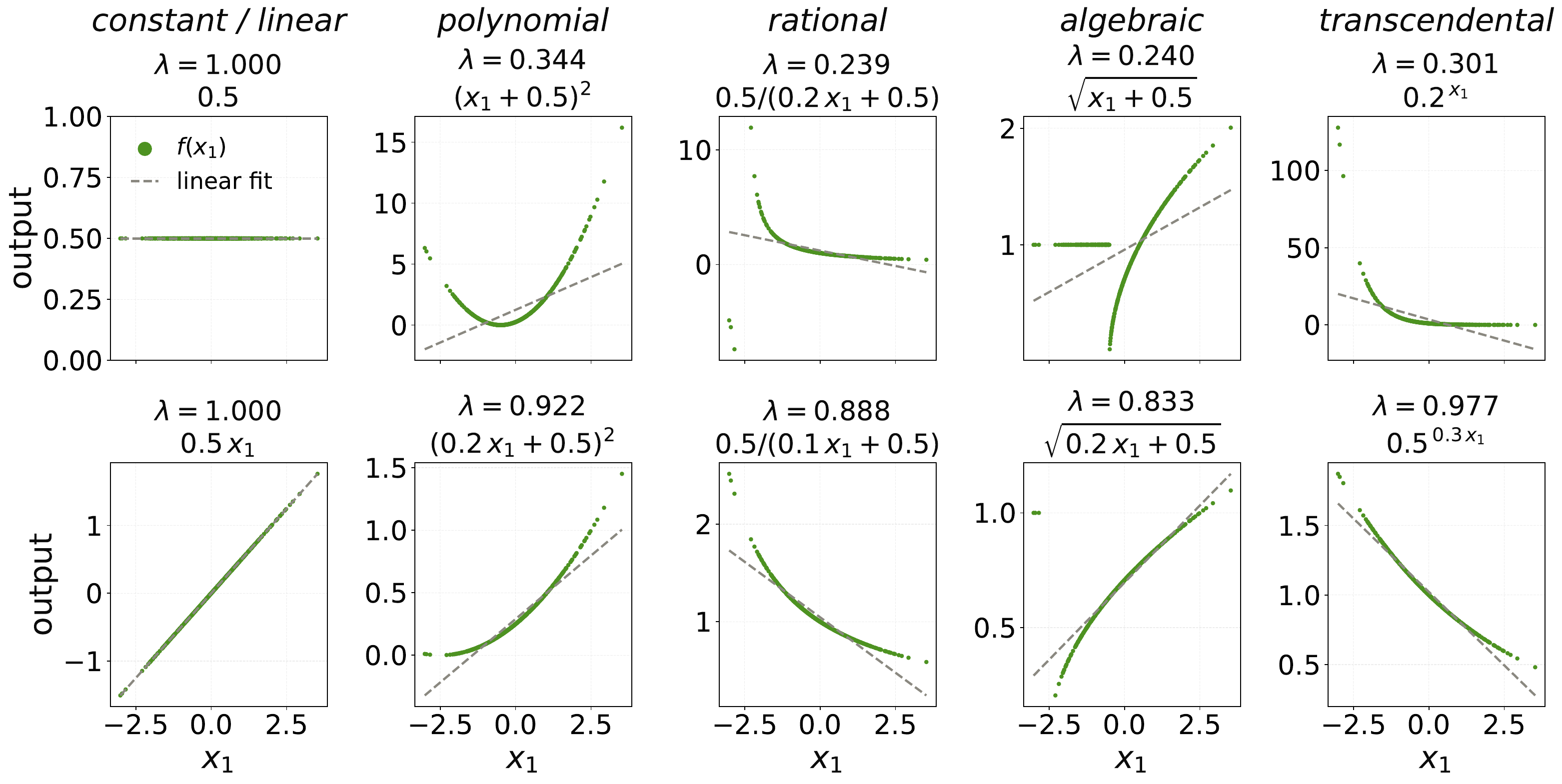}
\caption{Each plot shows the output of an example function on standardized inputs together with the least-squares linear fit. $\lambda$ is the adjusted $R^2$ of that fit. Each column shows two examples of one syntactic class. Constant and linear functions always have $\lambda = 1$. Although the non-linear examples differ only in their constants, the example functions in the bottom row behave more linearly (higher $\lambda$). As in the experiments, division and power are protected and return $1$ for undefined results (Table~\ref{tab:gp_parameters}).}
\label{fig:lambda-calibration-1d}
\end{figure}

Since solutions in SR are represented as programs in symbolic form, a naive way to measure their linearity would be to check whether a program uses only linear operations. However, this approach fails for syntactically non-linear programs as such programs can still behave nearly linearly on the inputs on which they are evaluated. For example, $(0.2\,x_1+0.5)^2$ is a non-linear polynomial, which behaves almost like a linear function within the considered input range (see Fig.~\ref{fig:lambda-calibration-1d}).

Therefore, we measure the linearity of a solution by its semantics instead of its syntax. Following the usual convention in semantic GP~\cite{krawiec2009approximating,moraglio2012gsgp,vanneschi2014semantic}, the \emph{semantics} of a program $f$ is the vector of its outputs $s(f) = \bigl(f(\mathbf{x}_1), \ldots, f(\mathbf{x}_m)\bigr) \in \mathbb{R}^m$ on a fixed set of inputs $X = [\mathbf{x}_1, \ldots, \mathbf{x}_m]$, where each $\mathbf{x}_i \in \mathbb{R}^d$ and $d$ is the dimensionality of the problem. The linearity of $s(f)$ is quantified by an ordinary least-squares regression of $s(f)$ on $X$. We define $\lambda(f)$ as its adjusted $R^2$, which corrects $R^2$ for the number of inputs $d$. This corresponds to an approach often used to characterize the linearity of a fitness landscape~\cite{mersmann2011ela}.

For $\lambda$ close to $1$, $f$ behaves linearly on $X$. The lower $\lambda$, the less of the behavior of $f$ a linear function explains. For the subsequent analysis, we distinguish between four linearity levels. We call a program \emph{linear} if $\lambda \geq 0.99$, \textit{nearly linear} if $0.9 \leq \lambda < 0.99$, \textit{moderately non-linear} if $0.5 \leq \lambda < 0.9$ and \textit{strongly non-linear} if $\lambda < 0.5$, meaning a linear fit explains less than half of the variance. Since $s(f)$, and thus $\lambda$, depends on the inputs $X$, the same program can behave almost linearly on a narrow input range and clearly non-linearly on a wide one. Throughout this work, we therefore compute $\lambda$ on standardized inputs with mean $0$ and standard deviation $1$, as used in the GP searches (Sections~\ref{sec:Generator} and~\ref{sec:Settings}).

Fig.~\ref{fig:lambda-calibration-1d} gives some examples with discrepancies between the syntax class and the semantics of a program. It lists $10$ one-dimensional example functions from five syntactic classes, ranging from constant or linear through polynomial, rational, and algebraic to transcendental. For the constant and linear example functions, the choice of constants has no effect on $\lambda$. In contrast, for the examples from non-linear classes, two functions with the same skeleton but different constants can differ strongly in $\lambda$. For example, the program $(x_1+0.5)^2$ has $\lambda = 0.34$, whereas $(0.2\,x_1+0.5)^2$ has $\lambda = 0.92$, and $0.2^{x_1}$ has $\lambda = 0.30$, whereas $0.5^{0.3\,x_1}$ has $\lambda = 0.98$. The examples illustrate that the linearity of a program can only be assessed by looking at its semantics and not its syntax. Thus, $\lambda$ is a proper measure for assessing the linearity of solutions as well as training data or offspring generated by search operators.

\subsection{Linearity of Benchmark Problems}
\label{sec:Benchmarks}

\begin{table}[t]
\centering
\caption{The synthetic and real-world benchmark problems, ordered by their exploitable non-linearity $R^2_{\mathrm{NN}}-\lambda$. $d$ is the dimensionality and $n$ the number of samples. $\lambda$ and $R^2_{\mathrm{NN}}$ are the held-out $R^2$ of a linear model and of a neural network fitted to the target.}
\label{tab:datasets}
\footnotesize
\begin{tabular}{llrrrrr}
\toprule
\textbf{Problem} & \textbf{Ground truth} & $d$ & $n$ & $\lambda$ & $R^2_{\mathrm{NN}}$ & $R^2_{\mathrm{NN}}-\lambda$\\
\midrule
\multicolumn{7}{l}{\textit{Feynman equations}} \\
I\_8\_14 & $\sqrt{(x_2-x_1)^2+(y_2-y_1)^2}$ & 4 & 10{,}000 & 0.00 & 1.00 & 1.00 \\
III\_17\_37 & $\beta(1+\alpha\cos\theta)$ & 3 & 10{,}000 & 0.13 & 1.00 & 0.87 \\
I\_50\_26 & $x_1(\cos\omega t+\alpha\cos^2\omega t)$ & 4 & 10{,}000 & 0.34 & 1.00 & 0.66 \\
II\_13\_17 & $2I/(4\pi\epsilon c^2 r)$ & 4 & 10{,}000 & 0.47 & 1.00 & 0.53 \\
II\_2\_42 & $\kappa(T_2-T_1)A/d$ & 5 & 10{,}000 & 0.60 & 1.00 & 0.40 \\
II\_34\_29b & $g\,\mu B J_z/(h/2\pi)$ & 5 & 10{,}000 & 0.68 & 1.00 & 0.32 \\
III\_4\_32 & $1/(e^{(h/2\pi)\omega/(k_bT)}-1)$ & 4 & 10{,}000 & 0.69 & 1.00 & 0.31 \\
II\_4\_23 & $q/(4\pi\epsilon r)$ & 3 & 10{,}000 & 0.72 & 1.00 & 0.28 \\
II\_10\_9 & $\sigma/(\epsilon(1+\chi))$ & 3 & 10{,}000 & 0.79 & 1.00 & 0.21 \\
II\_11\_3 & $qE_f/(m(\omega_0^2-\omega^2))$ & 5 & 10{,}000 & 0.82 & 1.00 & 0.18 \\
I\_39\_11 & $pV/(\gamma-1)$ & 3 & 10{,}000 & 0.83 & 1.00 & 0.17 \\
I\_25\_13 & $q/C$ & 2 & 10{,}000 & 0.83 & 1.00 & 0.17 \\
II\_8\_31 & $\epsilon E_f^2/2$ & 2 & 10{,}000 & 0.88 & 1.00 & 0.12 \\
II\_27\_18 & $\epsilon E_f^2$ & 2 & 10{,}000 & 0.88 & 1.00 & 0.12 \\
I\_34\_27 & $(h/2\pi)\,\omega$ & 2 & 10{,}000 & 0.93 & 1.00 & 0.07 \\
test\_7 & $\sqrt{8\pi G\rho/3 - \alpha c^2/d^2}$ & 5 & 10{,}000 & 0.98 & 1.00 & 0.02 \\
\midrule
\multicolumn{7}{l}{\textit{Real-world datasets}} \\
analcatdata\_apnea1 & -- & 3 & 475 & 0.01 & 0.89 & 0.88 \\
analcatdata\_apnea2 & -- & 3 & 475 & 0.01 & 0.82 & 0.81 \\
nikuradse\_1 & -- & 2 & 362 & 0.62 & 1.00 & 0.37 \\
visualizing\_galaxy & -- & 4 & 323 & 0.90 & 0.96 & 0.07 \\
rabe\_266 & -- & 2 & 120 & 0.98 & 1.00 & 0.02 \\
ESL & -- & 4 & 488 & 0.83 & 0.84 & 0.01 \\
LEV & -- & 4 & 1{,}000 & 0.52 & 0.52 & 0.00 \\
analcatdata\_neavote & -- & 2 & 100 & 0.94 & 0.94 & 0.00 \\
vinnie & -- & 2 & 380 & 0.77 & 0.76 & -0.01 \\
ERA & -- & 4 & 1{,}000 & 0.41 & 0.40 & -0.01 \\
visualizing\_environmental & -- & 3 & 111 & 0.26 & 0.24 & -0.02 \\
chscase\_geyser1 & -- & 2 & 222 & 0.76 & 0.73 & -0.03 \\
cloud & -- & 5 & 108 & 0.76 & 0.68 & -0.08 \\
\bottomrule
\end{tabular}
\end{table}
We now analyze the linearity of common SR benchmark problems.
In particular, we consider synthetic and real-world problems of PMLB~\cite{olson2017pmlb}, the collection that underlies SRBench~\cite{lacava2021srbench}, with $d=2$ to $d=5$ input variables, which is the range covered by TSGP in this work (Section~\ref{sec:Generator}). We take all $13$ real-world regression datasets with unknown ground truth with at least $100$ samples, where $n$ denotes the number of samples of a dataset. In addition, we use the Feynman equations~\cite{udrescu2020feynman}, whose targets follow a known physical formula. As the Feynman equations define many test problems, we draw $4$ per dimensionality at random, which gives us $16$ problems. Feynman datasets are randomly subsampled to $n = 10{,}000$ for computational efficiency. Table~\ref{tab:datasets} lists all $29$ problems.

To measure the linearity of a problem, we apply $\lambda$ to its target values. Unlike the semantics of a program, however, the target of a benchmark problem can contain noise, so a low $\lambda$ does not tell whether the variance the linear fit leaves unexplained is non-linear structure or noise. We therefore also fit a neural network (NN) with two hidden layers of $128$ and $64$ units. Since an NN can approximate any continuous function~\cite{hornik1989multilayer}, the variance it cannot explain comes from noise, insufficient training, or limited capacity. The difference $R^2_{\mathrm{NN}}-\lambda$ is the variance the NN explains beyond the linear model ($0$: no additional structure; close to $1$: almost entirely non-linear), which we call the \emph{exploitable non-linearity} of a problem. The remaining $1-R^2_{\mathrm{NN}}$ is the variance neither model explains, which usually comes from noise.

Since an in-sample evaluation would credit the NN for fitting the noise, we evaluate both models on held-out data. We fit them on four random $75/25$ splits and report the median. In each split, inputs and target are standardized to mean $0$ and standard deviation $1$ with the statistics of the training split. For problems, $\lambda$ is therefore the $R^2$ of the linear model on the test split. On held-out data, both values can be negative if a model predicts worse than the mean of the held-out targets. A problem is perfectly linear if $\lambda = 1$ and has no linear structure if $\lambda \leq 0$.

Both the Feynman and the real-world problems cover a wide range of $\lambda$. Among the Feynman problems, $\lambda$ ranges from $0.98$ for test\_7 to $0.00$ for I\_8\_14, and among the real-world problems from $0.98$ for rabe\_266 to $0.01$ for the two apnea datasets.

However, the two problem sets differ heavily in their exploitable non-linearity. On every Feynman problem, the NN explains the target perfectly ($R^2_{\mathrm{NN}} = 1.00$), which shows that the model's training and capacity are sufficient for these noise-free targets. All variance the linear model leaves unexplained is therefore exploitable. On the real-world problems, $R^2_{\mathrm{NN}}$ ranges from $1.00$ for rabe\_266 and nikuradse\_1 to $0.24$ for visualizing\_environmental. Only four of the $13$ problems show a clear exploitable non-linearity: the two apnea datasets ($0.88$ and $0.81$), nikuradse\_1 ($0.37$) and visualizing\_galaxy ($0.07$). On rabe\_266 and ESL, it is marginal ($0.02$ and $0.01$), and on the remaining seven problems it lies between $-0.08$ (cloud) and $0.00$. On these problems, the NN does not predict better than the linear model.

To sum up, many Feynman problems have a substantial exploitable non-linearity, whereas there are only a few real-world problems with non-linear structures. An operator intended to solve a wide range of problems should therefore be able to generate programs with both linear and non-linear structures.

\section{TSGP-2}
\label{sec:TSGP}
The training data of the original TSGP~\cite{anthes2025transformer,anthes2026dimensional} consists of programs generated by GP searches for linear target functions. As this might hurt search performance due to a potential linearity bias (Section~\ref{sec:Benchmarks}), we propose TSGP-2. TSGP-2 follows the model-building pipeline of TSGP (Section~\ref{sec:TSGP_Appraoch}), but generates both linear and non-linear target functions (Section~\ref{sec:Generator}). Section~\ref{sec:ArchiveLinearity} studies whether this leads to less linear training data.

\subsection{The TSGP Approach}\label{sec:TSGP_Appraoch}
TSGP is a model-based GP approach whose variation operator is a trained model that proposes, for a given program, another program with similar semantics. The model is trained once on separate training data (model building) and can then serve as variation operator for a standard GP search on different problems (model inference).

\paragraph{Model Building}
Training a model to propose semantic variations requires examples of program pairs that are semantically similar. Since the programs generated along a single GP run are variants of one another, TSGP obtains its training pairs from standard GP runs on synthetic problems~\cite{anthes2025transformer}. 
The pipeline is identical for TSGP and TSGP-2 and has five stages. For more details on the pipeline, we refer to the original work~\cite{anthes2025transformer}:

\begin{enumerate}
\item In a first step, synthetic datasets are generated for the GP searches. TSGP draws a random target function of dimensionality $d$ from a function generator (Section~\ref{sec:Generator}).\footnote{To ensure higher comparability, TSGP uses the same function generator as TSGP-2, restricted to the operator $\{+\}$ so that it produces only linear targets. The reported TSGP results are similar to those published in Anthes et al.~\cite{anthes2026dimensional}.} From that function, a synthetic regression dataset is created by drawing random inputs, computing their outputs and adding Gaussian noise to them. 
\item A standard GP run searches for a program that fits this dataset. Every program $f_i$ that is generated along the run is saved, together with its semantics $s(f_i)$, in an archive of that run. The Gaussian noise added to the function prevents the GP run from converging too fast and thus broadens the set of found programs.
\item The archived programs are normalized into a canonical form with SymPy~\cite{meurer2017sympy}. This removes syntactically distinct but semantically equivalent duplicates and sharpens the signal the model receives about which syntactic changes actually alter the semantics. 
\item Each archived program is paired with semantically similar programs, which are selected based on their \emph{semantic distance} ($\mathrm{SD}$), i.e., the Euclidean distance between their semantic vectors~\cite{moraglio2012gsgp}
\begin{equation}
  \mathrm{SD}(f_i, f_j) \;=\; \lVert s(f_i) - s(f_j) \rVert_2.
\end{equation}
As a result, each program $f_i$ forms pairs with its $k$ nearest neighbors $f_o \in N_k(f_i)$ within the same archive. Steps one to four are repeated several times, creating a set of archives with pairs of semantically similar programs. 
\item An encoder-decoder transformer~\cite{vaswani2017attention} is trained from scratch on the pairs of all archives to generate $f_o$ given $f_i$. In addition to $f_i$, the transformer also receives two conditioning values: the semantic distance $\mathrm{SD}(f_i, f_o)$ of the pair and the dimensionality $d$ of the problem the pair was drawn from. $\mathrm{SD}$ provides the model with a learning signal of how semantically similar two programs are, and $d$ conditions the model to propose programs that use at most $d$ variables.
\end{enumerate}

\paragraph{Model Inference}
Once trained, the model serves as the variation operator during search. For each parent, the conditioning values are set to a target semantic distance $\mathrm{SD}_t$ and the dimensionality $d$ of the problem. $\mathrm{SD}_t$ controls the desired semantic distance between parent and offspring (step size). Decoding is constrained by syntax control~\cite{wittenberg2022syntax}, which at each step assigns zero probability to every token that would make the sequence infeasible. Consequently, the operator always proposes a syntactically valid program.

\subsection{TSGP versus TSGP-2: The Target Functions Make the Difference}
\label{sec:Generator}
In this work, TSGP as well as TSGP-2 generate their target functions using a random expression generator based on the probabilistic tree sampler of~\cite{lample2019deep,biggio2021nesymres}. For a target function of dimensionality $d$, the generator first draws the number of operators, which is at least the number needed to combine all $d$ variables and at most two more. It then samples the tree shape, determining where unary and binary operators are placed~\cite{lample2019deep}. The leaves are filled with the input variables, drawn without replacement until each variable appears once and uniformly thereafter, so that every target depends on all $d$ variables. Each variable leaf is then wrapped as $a\,x+b$, with $a$ and $b$ taken from a fixed set of constants $\{0, 0.1, \ldots, 1\}$ and $a \neq 0$. 

The difference between TSGP and TSGP-2 lies in the set from which the operators are drawn: TSGP draws its operators only from $\{+\}$, so all target functions are linear by design, as in \cite{anthes2025transformer,anthes2026dimensional}. In contrast, TSGP-2 uses a larger operator set and draws the operators of the target function from the set $\{+, -, \times, \div, \sin, \sqrt{\cdot}\}$ with equal weights.

From each target function, a synthetic regression problem is derived, which serves as objective for a GP search. In particular, the function is evaluated on $1{,}000$ input points sampled uniformly from $[1,5]$; the corresponding outputs are perturbed with Gaussian noise of $0.1$ times their standard deviation. Then, both inputs and outputs are standardized. For each dimensionality $d \in \{2,3,4,5\}$, we generate $100$ problems and run one standard GP search on each, with a population size of $2{,}000$, until $100{,}000$ unique programs are archived or $300$ generations are reached. 

We want to emphasize that for both variants, TSGP and TSGP-2, the GP searching for solutions uses the same primitive set $\{+,-,\times,\%,\text{pow}\}$. Thus, neither model can propose a primitive the other cannot, and any difference between their training data, and between the learned operators, lies only in how these primitives are combined by the GP to solve the different target functions.

Each archived program is paired with its $k=3$ nearest neighbors. From the resulting pairs, we draw $2.5$ million pairs per dimensionality, resulting in $10$ million pairs in total, and hold out $10\%$ of them for validation. This is half as many pairs as in~\cite{anthes2026dimensional}, whose ablation studies showed that the training budget can be reduced in the number of pairs without a significant loss in performance. We train the TSGP and TSGP-2 models for $8$ epochs. The remaining pairs, which are not drawn for training or validation, form a separate test set.

\subsection{Linearity of the Training Data}
\label{sec:ArchiveLinearity}
We analyze whether the non-linear targets used for TSGP-2 indeed lead to less linear training data. This is not guaranteed, since the targets enter the training data only indirectly through the programs GP generates while searching for them. To analyze the linearity of the training data generated by the GP, we randomly draw $50{,}000$ of the $10$ million training pairs of each model. For each pair, we compute $\lambda$ (Section~\ref{sec:Lambda}) for both programs on $500$ input points drawn from a standard normal distribution, resulting in $100{,}000$ evaluated programs per model.

\begin{figure}[t]
\centering
\includegraphics[width=0.72\linewidth]{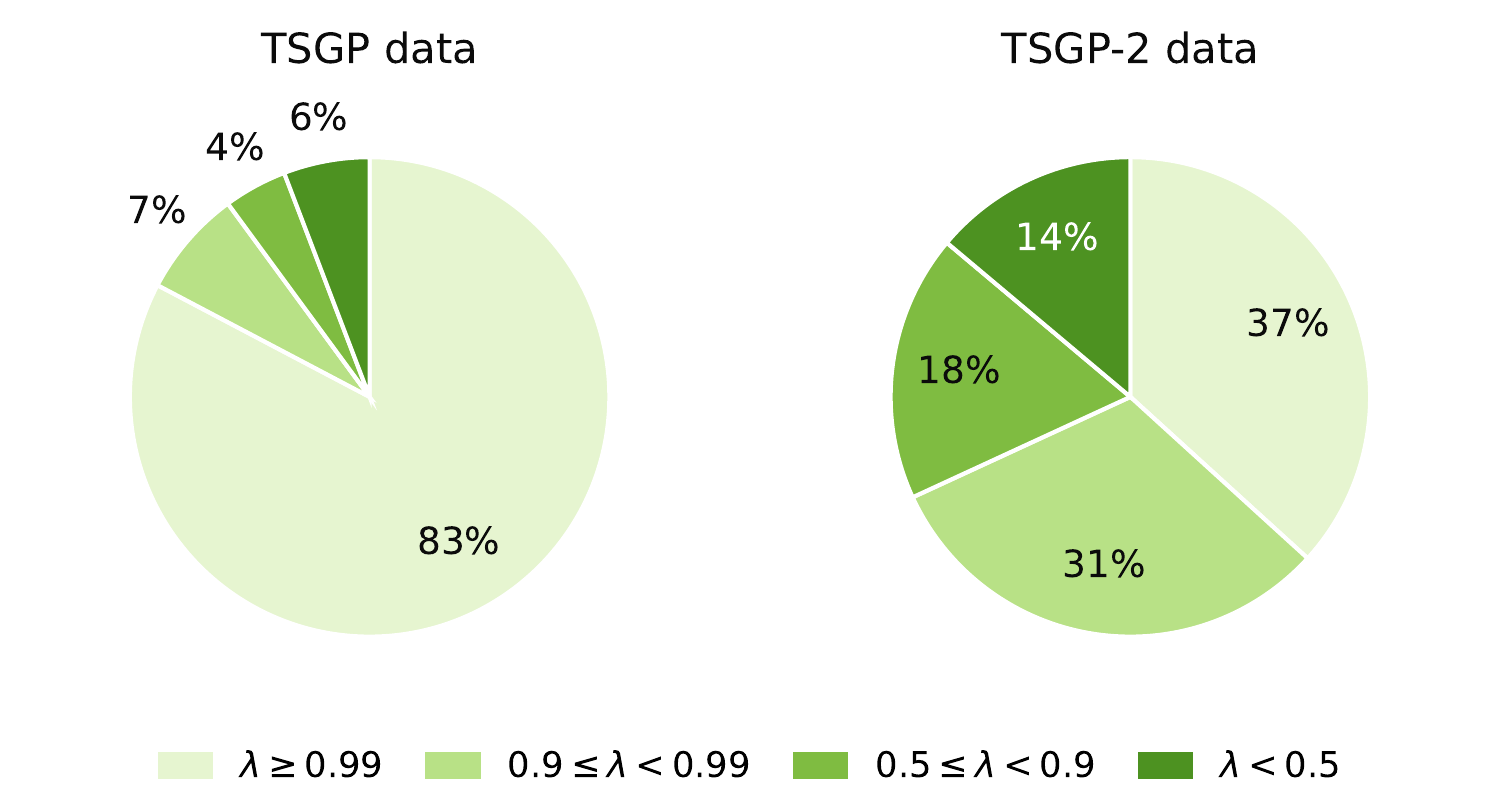}
\caption{Share of each linearity level among $100{,}000$ sampled training programs of TSGP (left) and TSGP-2 (right).}
\label{fig:archive-lambda-pie}
\end{figure}

Fig.~\ref{fig:archive-lambda-pie} compares the linearity of the training data for TSGP (left) and TSGP-2 (right). We find that the linearity of the targets carries over to the training data. In the TSGP training data, $82.7\%$ of the programs behave linearly, while only $7.2\%$, $4.3\%$, and $5.8\%$ of the programs are nearly linear, moderately non-linear, and strongly non-linear, respectively. The situation is different for TSGP-2 training data, where we observe a lower share of programs that behave linearly. Only $36.8\%$ of the programs behave linearly, while $31.4\%$ are nearly linear, $18.0\%$ moderately and $13.9\%$ strongly non-linear. Thus, non-linear behavior of the programs is far more frequent than in the TSGP data: about a third of the TSGP-2 programs are moderately or strongly non-linear, compared to only a tenth of the TSGP programs. Thus, the training data of TSGP is strongly biased toward linear behavior, whereas the use of non-linear targets in TSGP-2 reduces this bias considerably. The following Section~\ref{sec:Search} examines whether this difference improves search performance, and Section~\ref{sec:Analysis} analyzes the behavior of the variation operator. 

\section{Search Performance}\label{sec:Search}
To analyze the effect of the training data on solution quality, we compare the performance of TSGP-2 to TSGP and standard GP.\footnote{The code and trained model weights are available here: \url{https://github.com/Giganth/transformer_semantic_gp}.}

\subsection{Experimental Setting}
\label{sec:Settings}

All three methods (TSGP, TSGP-2, and standard GP) use the parameters in Table~\ref{tab:gp_parameters} and differ only in their variation operator. Each search evolves a population of $100$ programs over $100$ generations. The initial population is created with ramped half-and-half and depths from $2$ to $5$, and
parents are selected by tournament selection of size $5$ without elitism. The primitive set is the same as in model building. It consists of the problem variables, ephemeral random constants (ERCs) from $-0.5$ to $0.5$ in steps of $0.1$, and the operators $+$,~$-$,~$\times$,~protected division (\%) and protected power (pow), which return $1$ for undefined results.

Standard GP applies subtree crossover with a probability of $0.9$ and subtree mutation with a probability of $0.1$, which inserts a new subtree of depth $0$ to $2$. Following Koza~\cite{koza1992gp}, the maximum tree depth is $17$. TSGP and TSGP-2 generate one offspring per selected parent, both with a target semantic distance of $\mathrm{SD}_t = 1$. 

For each run, we split the data of a problem at random into $75\%$ training and $25\%$ test cases and standardize inputs and target with the mean and standard deviation of the training cases. As the solution of a run, we report the program with the lowest training RMSE over all generations and evaluate it on the test cases. On each problem, we compare every pair of methods by their test errors over the $30$ runs with a two-sided Mann-Whitney-U-test. Within each problem set, we rank the methods on each problem by their median test error and compare the ranks of every pair with a Wilcoxon signed-rank test. Both tests are Holm-corrected over the three pairwise comparisons at a significance level of $0.05$.

\begin{table}[t]
\centering
\caption{Parameters of the searches. $\%$ and pow are protected operators, which return $1$ for undefined results.}
\label{tab:gp_parameters}
\footnotesize
\begin{tabular}{ll}
\toprule
\textbf{Parameter} & \textbf{Value} \\
\midrule
Population size & $100$ \\
Generations & $100$ \\
Initialization & Ramped half-and-half, depth $[2, 5]$ \\
Selection & Tournament, size $5$ \\
Primitive set & $\{x_1, \dots, x_d, \text{ERC}, +, -, \times, \%, \text{pow}\}$ \\
ERC & $[-0.5, 0.5]$, step size $0.1$ \\
Fitness & RMSE on the training cases \\
Data split & $75/25$ training/test \\
Runs & $30$ per method and problem \\
\bottomrule
\end{tabular}
\end{table}

\subsection{Solution Quality}
\label{sec:Quality}
We test whether the difference in constructing the training data between TSGP and TSGP-2 affects solution quality, and whether this effect depends on the exploitable non-linearity of a problem (compare Table~\ref{tab:datasets}).

\begin{table}[!t]
\centering
\caption{Median test RMSE of the solutions over $30$ runs, with the interquartile range over the runs in parentheses. Lower is better and the best method per problem is bold. For each problem, superscripts indicate all other methods that are significantly worse (Mann-Whitney per problem and Wilcoxon signed-rank over the per-problem ranks; both are Holm-corrected). Problems are ordered as in Table~\ref{tab:datasets}, from the highest exploitable non-linearity to the lowest.}
\label{tab:performance}
\footnotesize
\setlength{\tabcolsep}{4pt}
\begin{tabular}{lrrr}
\toprule
\textbf{Problem} & \textbf{Standard GP}~\textit{(a)} & \textbf{TSGP}~\textit{(b)} & \textbf{TSGP-2}~\textit{(c)} \\
\midrule
\multicolumn{4}{l}{\textit{Feynman equations}} \\
I\_8\_14 & 0.774\,{\scriptsize(0.096)} & $^{a}$\,0.696\,{\scriptsize(0.122)} & $^{a}$\,\textbf{0.633}\,{\scriptsize(0.223)} \\
III\_17\_37 & 0.578\,{\scriptsize(0.187)} & 0.500\,{\scriptsize(0.135)} & \textbf{0.488}\,{\scriptsize(0.336)} \\
I\_50\_26 & 0.803\,{\scriptsize(0.037)} & 0.793\,{\scriptsize(0.020)} & \textbf{0.792}\,{\scriptsize(0.022)} \\
II\_13\_17 & 0.543\,{\scriptsize(0.263)} & $^{a}$\,\textbf{0.470}\,{\scriptsize(0.111)} & $^{a}$\,0.484\,{\scriptsize(0.114)} \\
II\_2\_42 & 0.622\,{\scriptsize(0.178)} & $^{a}$\,0.402\,{\scriptsize(0.214)} & $^{a}$\,\textbf{0.326}\,{\scriptsize(0.173)} \\
II\_34\_29b & 0.503\,{\scriptsize(0.155)} & $^{a}$\,0.339\,{\scriptsize(0.089)} & $^{ab}$\,\textbf{0.312}\,{\scriptsize(0.079)} \\
III\_4\_32 & 0.451\,{\scriptsize(0.250)} & $^{a}$\,0.359\,{\scriptsize(0.096)} & $^{ab}$\,\textbf{0.314}\,{\scriptsize(0.076)} \\
II\_4\_23 & 0.394\,{\scriptsize(0.242)} & $^{a}$\,0.318\,{\scriptsize(0.060)} & $^{ab}$\,\textbf{0.272}\,{\scriptsize(0.104)} \\
II\_10\_9 & 0.466\,{\scriptsize(0.301)} & $^{a}$\,0.256\,{\scriptsize(0.070)} & $^{ab}$\,\textbf{0.234}\,{\scriptsize(0.066)} \\
II\_11\_3 & 0.468\,{\scriptsize(0.240)} & $^{a}$\,0.288\,{\scriptsize(0.080)} & $^{ab}$\,\textbf{0.250}\,{\scriptsize(0.048)} \\
I\_39\_11 & 0.267\,{\scriptsize(0.221)} & $^{a}$\,0.190\,{\scriptsize(0.050)} & $^{ab}$\,\textbf{0.140}\,{\scriptsize(0.058)} \\
I\_25\_13 & 0.175\,{\scriptsize(0.171)} & 0.186\,{\scriptsize(0.038)} & \textbf{0.160}\,{\scriptsize(0.091)} \\
II\_8\_31 & 0.264\,{\scriptsize(0.221)} & $^{a}$\,\textbf{0.104}\,{\scriptsize(0.033)} & $^{a}$\,\textbf{0.104}\,{\scriptsize(0.042)} \\
II\_27\_18 & 0.240\,{\scriptsize(0.260)} & $^{a}$\,0.103\,{\scriptsize(0.026)} & $^{a}$\,\textbf{0.099}\,{\scriptsize(0.033)} \\
I\_34\_27 & 0.186\,{\scriptsize(0.233)} & $^{a}$\,\textbf{0.015}\,{\scriptsize(0.013)} & $^{a}$\,\textbf{0.015}\,{\scriptsize(0.006)} \\
test\_7 & 0.152\,{\scriptsize(0.336)} & $^{a}$\,0.120\,{\scriptsize(0.030)} & $^{ab}$\,\textbf{0.097}\,{\scriptsize(0.008)} \\
\cmidrule(l){1-4}
\textit{mean rank} & 2.94 & $^{a}$\,1.88 & $^{ab}$\,1.19 \\
\midrule
\multicolumn{4}{l}{\textit{Real-world datasets}} \\
analcatdata\_apnea1 & \textbf{0.532}\,{\scriptsize(0.600)} & 0.855\,{\scriptsize(0.385)} & 0.737\,{\scriptsize(0.527)} \\
analcatdata\_apnea2 & \textbf{0.541}\,{\scriptsize(0.424)} & 0.845\,{\scriptsize(0.331)} & 0.635\,{\scriptsize(0.567)} \\
nikuradse\_1 & 0.385\,{\scriptsize(0.205)} & 0.373\,{\scriptsize(0.107)} & \textbf{0.333}\,{\scriptsize(0.125)} \\
visualizing\_galaxy & 0.312\,{\scriptsize(0.087)} & $^{a}$\,0.284\,{\scriptsize(0.042)} & $^{ab}$\,\textbf{0.270}\,{\scriptsize(0.048)} \\
rabe\_266 & 0.235\,{\scriptsize(0.117)} & $^{a}$\,0.094\,{\scriptsize(0.026)} & $^{a}$\,\textbf{0.088}\,{\scriptsize(0.024)} \\
ESL & 0.497\,{\scriptsize(0.171)} & $^{a}$\,0.391\,{\scriptsize(0.049)} & $^{a}$\,\textbf{0.390}\,{\scriptsize(0.047)} \\
LEV & 0.722\,{\scriptsize(0.128)} & $^{a}$\,\textbf{0.675}\,{\scriptsize(0.038)} & $^{a}$\,0.679\,{\scriptsize(0.036)} \\
analcatdata\_neavote & 0.271\,{\scriptsize(0.082)} & \textbf{0.265}\,{\scriptsize(0.096)} & \textbf{0.265}\,{\scriptsize(0.116)} \\
vinnie & 0.525\,{\scriptsize(0.051)} & 0.523\,{\scriptsize(0.048)} & \textbf{0.520}\,{\scriptsize(0.044)} \\
ERA & 0.818\,{\scriptsize(0.064)} & $^{a}$\,\textbf{0.791}\,{\scriptsize(0.036)} & $^{a}$\,0.793\,{\scriptsize(0.048)} \\
visualizing\_environmental & 0.901\,{\scriptsize(0.184)} & 0.909\,{\scriptsize(0.214)} & \textbf{0.874}\,{\scriptsize(0.169)} \\
chscase\_geyser1 & 0.508\,{\scriptsize(0.076)} & \textbf{0.499}\,{\scriptsize(0.057)} & 0.501\,{\scriptsize(0.062)} \\
cloud & 0.635\,{\scriptsize(0.308)} & \textbf{0.523}\,{\scriptsize(0.289)} & 0.588\,{\scriptsize(0.316)} \\
\cmidrule(l){1-4}
\textit{mean rank} & 2.62 & 1.92 & $^{a}$\,1.46 \\
\bottomrule
\end{tabular}
\end{table}

Table~\ref{tab:performance} reports the median test error of the solutions over the $30$ runs of each problem.
On the Feynman problems, TSGP-2 achieves the best mean rank ($1.19$), which is significantly better than those of TSGP ($1.88$) and standard GP ($2.94$). Both TSGP variants significantly outperform standard GP on $13$ problems, and their errors vary considerably less between runs.
TSGP-2 finds the best solution on $13$ of the $16$ problems (and ties on two) and is significantly better than TSGP on $7$ of them, whereas TSGP is never significantly better than TSGP-2. The performance gains of TSGP-2 are mostly present for problems with medium or high exploitable non-linearity (Table~\ref{tab:datasets}). Six of the seven problems on which TSGP-2 is significantly better have an exploitable non-linearity between $0.17$ and $0.32$. For problems with very high linearity, where a linear fit explains most of the structure, TSGP and TSGP-2 reach almost identical errors. On III\_17\_37 and I\_50\_26, two of the most non-linear problems, no method is significantly better than another. We assume that this is due to the cosine structure of these problems, which is not part of the primitive set of the methods (Table~\ref{tab:gp_parameters}). 

On the real-world problems, TSGP-2 achieves a mean rank of $1.46$ and significantly outperforms standard GP ($2.62$) but not TSGP ($1.92$). This is expected, as most real-world problems have little exploitable non-linearity. However, TSGP-2 still achieves better results than TSGP for the four problems with exploitable non-linearity (for visualizing\_galaxy even significantly better). On the two apnea datasets, which carry the most exploitable non-linearity, standard GP finds the best solutions, but TSGP-2 reduces the error of TSGP from $0.86$ to $0.74$ and from $0.85$ to $0.64$. Due to the strong variation between runs (interquartile ranges of $0.33$ to $0.60$), however, this difference is not significant. On the nine problems without exploitable non-linearity, TSGP and TSGP-2 perform similarly. Both find the better solution on four problems (with one tie), and none of the differences is significant.

In summary, TSGP-2 achieves the best mean rank on both problem sets. TSGP-2 outperforms TSGP on problems with exploitable non-linearity, achieving significantly better results on $8$ of them. For problems without exploitable non-linearity, both achieve a similar solution quality.

\section{Analysis of the Linearity Bias}
\label{sec:Analysis}
Section~\ref{sec:Quality} shows that TSGP-2 finds better solutions than TSGP, mainly on problems with exploitable non-linearity. To understand where this advantage comes from, we analyze the linearity of the best program of each run (Section~\ref{sec:Linearity_Solutions}), the linearity of the offspring proposed and selected during the search (Section~\ref{sec:Linearity_Population}), and the linearity of the offspring created by the search operator (Section~\ref{sec:OperatorLinearity}).

\subsection{Error and Linearity of the Best Programs}
\label{sec:Linearity_Solutions}
\begin{figure}[t]
\centering
\includegraphics[width=\linewidth]{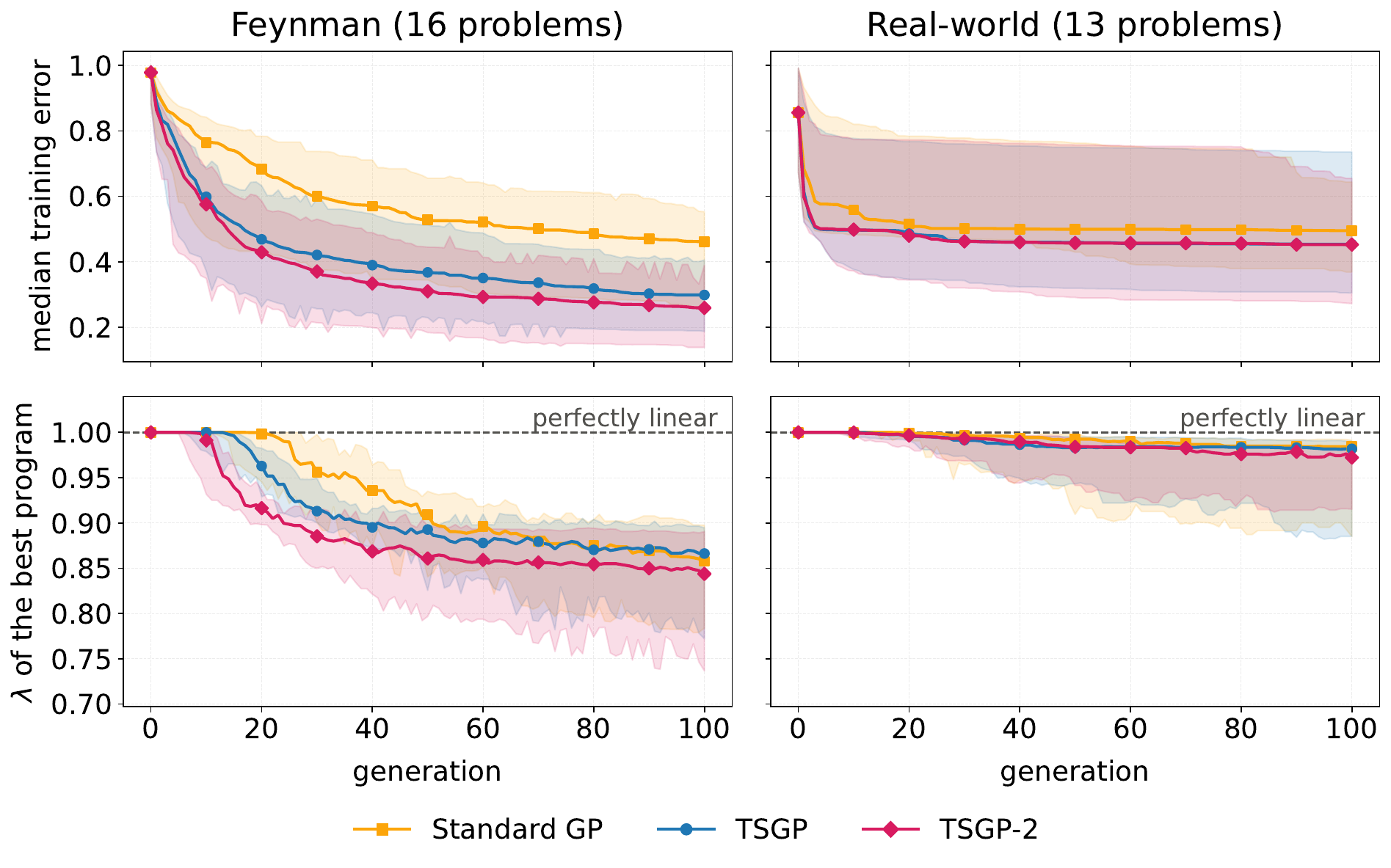}
\caption{Training error (top) and linearity $\lambda$ (bottom) of the best program over the run, for the Feynman (left) and real-world problems (right); we show the median over runs and problems. Bands are bootstrap intervals over problems.}
\label{fig:best-program}
\end{figure}

We analyze how the linearity and error of the best programs evolve over the search.
Fig.~\ref{fig:best-program} shows the training error (top) and the linearity $\lambda$ (bottom) of the \emph{best program} (i.e., the program with the lowest training error so far) over the generations. Linearity is measured on the training cases of each run. We plot the median over the $30$ runs for each problem and over all problems in each set (Feynman versus real-world). Appendix~\ref{app:per-problem} shows results for each problem.

On the Feynman problems, both TSGP variants reduce the error much faster than standard GP from the start, while they barely differ from each other in the first generations. From about generation $15$ on, the error difference increases. At generation $40$, the median error of TSGP-2 is $0.33$ compared to $0.39$ for TSGP, and it remains lower until the end of the run ($0.26$ compared to $0.30$).

The best programs of TSGP-2 also leave the linear region first. Starting from a median $\lambda$ of $1.00$ for all methods, the median $\lambda$ of TSGP-2 drops below $0.99$ at generation $11$, that of TSGP at generation $16$ and that of standard GP at generation $25$. This happens shortly before the error advantage of TSGP-2 over TSGP increases. After around $20$ generations, the difference in linearity is clearly visible ($\lambda=0.92$ for TSGP-2, $\lambda=0.96$ for TSGP and $\lambda=1.00$ for standard GP), before the difference in error becomes large. We conclude that TSGP takes longer than TSGP-2 to find good programs outside the linear region, and TSGP-2 keeps the resulting error advantage until the end of the search.

The comparison with standard GP shows that more non-linearity alone does not automatically lead to better solutions. From about generation $70$ on, the best programs of standard GP are as non-linear as those of TSGP ($0.86$ vs.\ $0.87$ at generation $100$), but their training error is considerably higher ($0.46$ vs.\ $0.30$). The non-linear programs that standard GP finds are worse than those of TSGP and TSGP-2. This indicates that the learned operators propose non-linear programs that are more relevant to the target.

For the real-world problems, both TSGP variants also reduce the error faster than standard GP, from $0.86$ to about $0.50$ within $10$ generations, whereas standard GP reaches $0.56$. Afterward, all three methods improve only slowly, and by the end of the run the TSGP variants reach a median error of $0.45$ and standard GP of $0.50$. For all three methods, the best programs found remain nearly linear throughout the run, with a median $\lambda$ above $0.97$. TSGP and TSGP-2 barely differ in error and linearity throughout the run. This is expected, since nine of the $13$ real-world problems carry no exploitable non-linearity. On the real-world problems with the highest exploitable non-linearity, the two apnea datasets, the best programs found by TSGP are much more linear than those of TSGP-2; TSGP-2 finds solutions with lower error, whereas TSGP reduces the error only slightly over a run (Figs.~\ref{fig:error-per-problem-realworld} and~\ref{fig:linearity-per-problem-realworld} in Appendix~\ref{app:per-problem}).

In summary, the best programs of TSGP-2 leave the linear region earlier than those of TSGP on problems with exploitable non-linearity. After they leave the linear region, the error advantage of TSGP-2 over TSGP grows, and both differences persist until the end of the search. On the Feynman problems, standard GP also finds non-linear solutions, but they have higher errors, indicating that the non-linearity induced by both TSGP variants is more beneficial for search performance.

\subsection{Linearity of Populations over the Number of Generations}
\label{sec:Linearity_Population}
\begin{figure}[t]
    \centering
    \includegraphics[width=\linewidth]{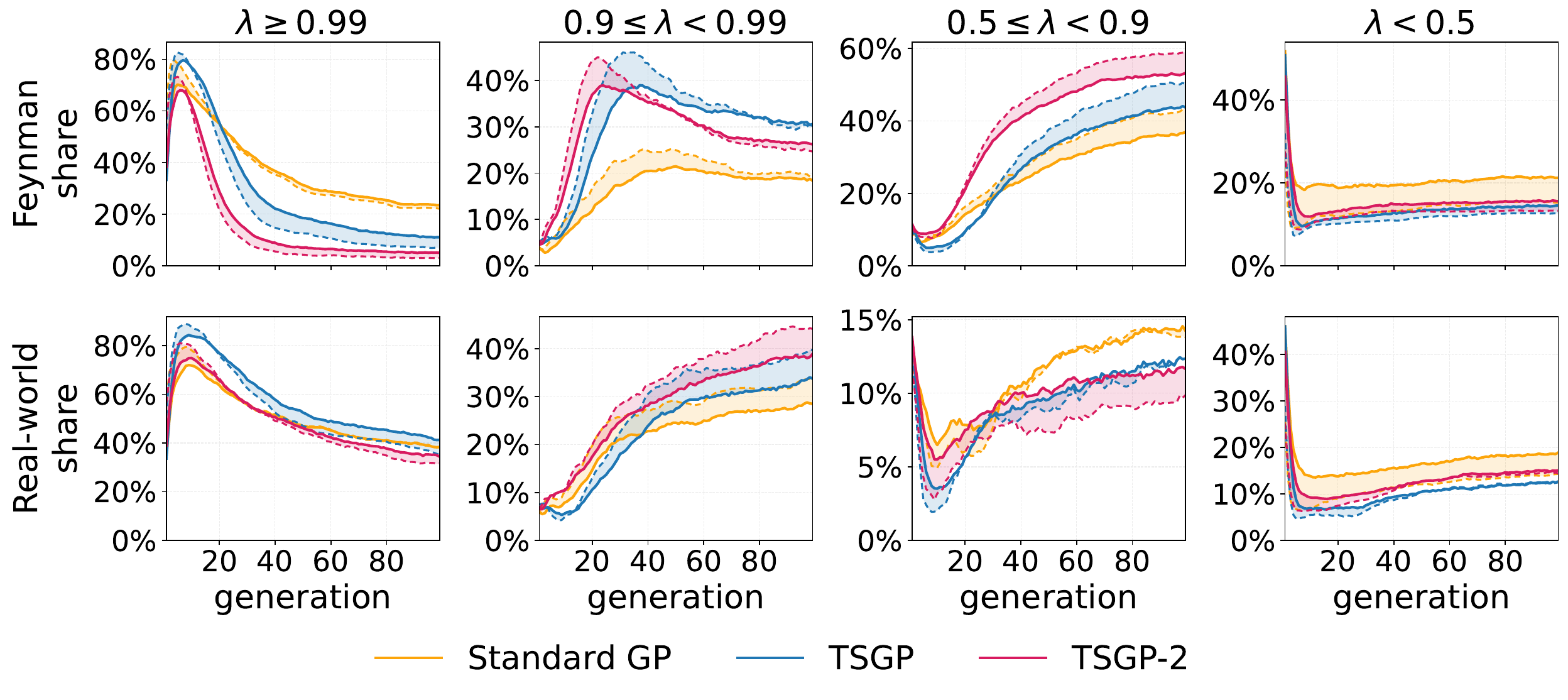}
    \caption{Percentage of offspring that belong to a linearity level over the number of generations. Results are for all offspring (solid) and selected offspring (dashed) and for Feynman (top) and real-world problems (bottom). 
    If a dashed line is above the corresponding solid one, selection favors solutions of this linearity level. If it is below, selection disfavors it, indicating low fitness.}
    \label{fig:selection}
\end{figure}

The previous subsection shows that TSGP-2 is able to find non-linear solutions with high performance faster than TSGP. To understand why, we study how the linearity of the created solutions changes over the number of generations. For the four different linearity levels (Section~\ref{sec:Lambda}), Fig.~\ref{fig:selection} plots the percentage of solutions that belong to a given linearity level over the number of generations. We show results for all offspring before selection (solid line) and after selection (dashed line). Results are for Feynman (top) and real-world problems (bottom). The percentages are averaged over all runs and problems. 

We start our analysis with the Feynman problems. At the start of the search, all three methods mainly generate linear solutions. Within a few generations, the share of linear offspring in a population reaches its maximum of about $70\%$ for standard GP and TSGP-2 and $80\%$ for TSGP (top, left plot). After this peak, TSGP-2 quickly reduces the number of linear solutions and at generation $20$ only $29\%$ of its offspring are still linear, compared with about $55\%$ for TSGP and standard GP. Although selection in TSGP clearly disfavors linear structures (there is a large gap between the blue solid and dashed line), TSGP populations still contain many more linear structures than those of TSGP-2. 

TSGP-2 is fastest in creating nearly linear structures (second column) as well as moderately non-linear structures (third column). For the nearly linear structures, TSGP-2 already reaches a maximum of $39\%$ of the population at generation $24$; TSGP needs $37$ generations until it reaches the same level, whereas standard GP stays below $21\%$. For all methods, the percentage of nearly linear programs (until about generation $50$) and of moderately non-linear programs (from about generation $25$ until the end of the search) is higher in the population after selection than before. Thus, these programs have higher fitness and are preferred by selection.

For the strongly non-linear structures (top, right), standard GP creates about $20\%$ strongly non-linear offspring in each generation throughout the search, but these make up only $11\%$ to $15\%$ of the selected offspring. For the TSGP variants, the percentage is lower and a higher share of these solutions is kept by selection. Thus, standard GP often proposes strongly non-linear offspring, but these solutions have low fitness and are often discarded by selection. In contrast, the two TSGP variants generate fewer non-linear structures; however, the created solutions usually have high fitness and survive selection.

For the real-world problems, the situation is different, since optimal solutions have a much more linear structure (compare Table~\ref{tab:datasets}). Over the number of generations, the percentage of linear offspring (bottom, left) remains much higher for all three search approaches. After a few generations, the number of nearly linear structures starts increasing until the end of the search. The number of moderately non-linear as well as strongly non-linear structures slightly increases over time; however, their absolute numbers remain low. For example, non-linear structures created by TSGP-2 usually have lower fitness and thus selection reduces the percentage of such structures in a population. This is different for nearly linear structures, which are favored by selection (dashed line is higher than the solid line). Thus, for the real-world problems, which have only little exploitable non-linearity, the creation of non-linear structures gives no advantage to the search, as those structures usually have lower fitness in comparison to more linear structures. These findings are consistent with the performance of the search methods (compare Section~\ref{sec:Quality}).

In summary, search usually starts with creating linear structures. As the Feynman problems have a higher level of exploitable non-linearity, search methods move toward creating more non-linear structures. TSGP-2 is fastest in moving from high-quality linear structures to high-quality non-linear structures. In contrast, standard GP creates many non-linear structures, but most of them have low quality and are discarded by selection. The situation is different for the real-world problems, which have much less exploitable non-linearity and can often be solved with more linear programs. For these problems, it is not relevant to generate high-quality non-linear structures. Thus, all three methods are able to create high-quality linear or nearly linear structures. 

\subsection{Analyzing the Linearity Bias of the Search Operator}
\label{sec:OperatorLinearity}

\begin{figure}[t]
    \centering
    \includegraphics[width=\linewidth]{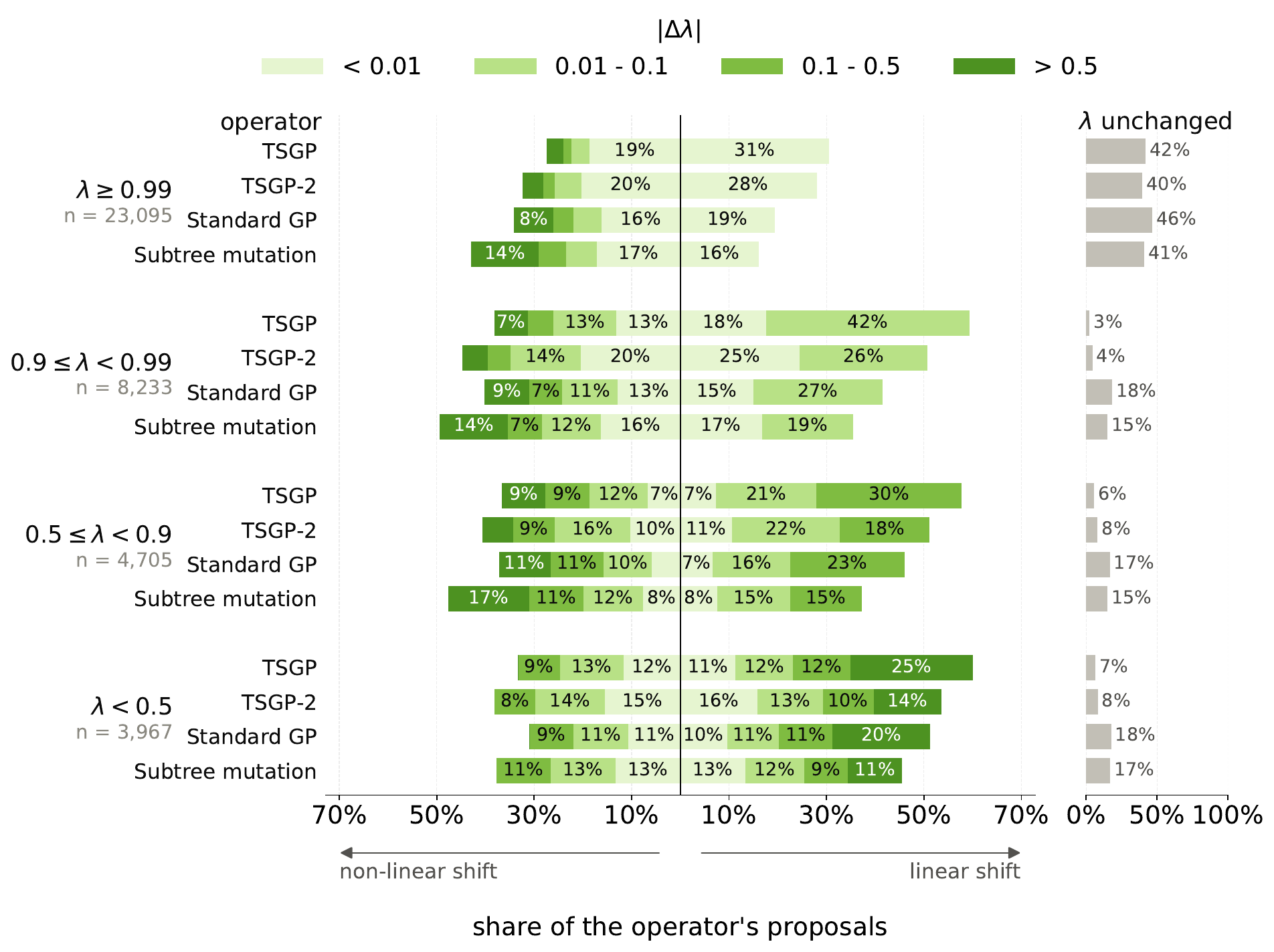} 
    \caption{Change in linearity $\Delta\lambda$ from parent to offspring, per linearity level of the parent and operator. Bars left of zero lower $\lambda$, bars right of zero raise it, and the shade indicates $|\Delta\lambda|$. The right panel shows the share of offspring with unchanged $\lambda$. All operators vary the same $40{,}000$ parents from test pairs held out from the TSGP and TSGP-2 training data.}
    \label{fig:transition-operator}
\end{figure}

The previous section studied how the linearity of populations changes over the number of generations. We found that the search of TSGP has a stronger linearity bias than TSGP-2 and is slower in generating high-quality non-linear solutions.
This section goes deeper and looks at the search operator. We want to understand whether the search operator has a bias toward creating more linear or non-linear offspring. 

Consequently, we apply the search operator to one parent and measure the change in linearity of the offspring. As parents, we use the input programs of held-out test pairs from the TSGP and TSGP-2 training data. We use only programs that are not included in any training or validation pair of any model, neither as input nor as output. From the test pairs of each model (TSGP and TSGP-2), we randomly draw $20{,}000$ parents. To study the bias of the different operators, we apply each operator once to each of these $40{,}000$ parents. TSGP as well as TSGP-2 are conditioned on $\mathrm{SD}_t = 1$. As baselines, we use the variation operators of standard GP with the settings described in Section~\ref{sec:Settings}.

For each parent, we measure the change in linearity $\Delta\lambda = \lambda(\text{offspring}) - \lambda(\text{parent})$. A positive $\Delta\lambda$ means the operator creates an offspring that is more linear than its parent; a negative $\Delta\lambda$ indicates a more non-linear offspring. Offspring with $|\Delta\lambda| \leq 10^{-6}$ count as unchanged. We report results separately for the four different linearity levels of the parents, since $\Delta\lambda$ depends on the linearity of the parent, e.g., if a parent is linear, it is not possible to obtain an offspring with $\Delta\lambda>0.01$.

For the four different linearity levels of the parent, Fig.~\ref{fig:transition-operator} shows $\Delta\lambda$ for the different types of operators (TSGP, TSGP-2, standard GP, and subtree mutation). Below the linearity level of the parent, we plot the number $n$ of parents that belong to this level. The vertical line separates all offspring that are more linear than their parents (linear shift) from the offspring that are more non-linear than their parents (non-linear shift). For example, the first row of Fig.~\ref{fig:transition-operator} shows that when applying the TSGP operator to the $23{,}095$ linear parents ($\lambda \geq 0.99$), $42\%$ of the offspring are unchanged ($|\Delta\lambda| \leq 10^{-6}$), $31\%$ are slightly more linear ($10^{-6} < \Delta\lambda < 0.01$), and $19\%$ are slightly more non-linear ($-0.01 < \Delta\lambda < -10^{-6}$).

For linear parents ($\lambda \geq 0.99$), the offspring remain mostly linear independently of the operator. TSGP-2 creates more structures with lower $\lambda(\text{offspring})$ in comparison to TSGP ($32\%$ versus $27\%$). TSGP as well as TSGP-2 lower $\lambda$ less often than standard GP and subtree mutation ($34\%$ and $43\%$). However, as we have seen in Section~\ref{sec:Linearity_Population}, the strongly non-linear offspring produced by standard GP often have low fitness and do not survive the selection step.

For all four linearity levels of the parents, we find that the TSGP operator has the strongest bias toward linear structures. For TSGP, $31\%$, $59\%$, $58\%$, and $60\%$ of the offspring are more linear than the parent for linear, nearly linear, moderately non-linear, and strongly non-linear parents, respectively. For TSGP-2, the linearity bias is lower ($28\%$, $51\%$, $51\%$, and $54\%$). TSGP not only moves offspring toward linearity more often, but also by larger steps, which is most visible for nearly linear parents. TSGP moves $42\%$ of them close to or onto the linear level, compared to only $26\%$ for TSGP-2, $27\%$ for standard GP, and $19\%$ for subtree mutation.

\section{Conclusions}\label{sec:Conclusion}
This work analyzed the linearity of SR benchmark problems and proposed TSGP-2, which reduces the linearity bias that TSGP inherits from its training data. 
While most real-world problems can be well solved by linear structures (although not all of them), solving Feynman problems requires non-linear structures. The training data of TSGP, however, consists mainly of programs with linear behavior.

We therefore proposed TSGP-2, which trains a model on data that consists of both linear and non-linear program behavior, reducing the share of linear programs from $83\%$ to $37\%$. We find that TSGP-2 achieves the best mean rank across the studied problems (Feynman and real-world) and clearly outperforms TSGP on problems with exploitable non-linearity. On the Feynman problems, TSGP-2 is significantly better on $7$ out of $16$ problems compared to TSGP, and never significantly worse. On the real-world problems, TSGP-2 is the only operator that significantly outperforms standard GP and improves especially on the problems with exploitable non-linearity. 

On the remaining problems without exploitable non-linearity, TSGP-2 and TSGP reach comparable solution quality. On these problems, selection filters out non-linear structures that are not relevant for the search. We conclude that the linearity bias mostly harms performance on problems where non-linear structures are needed and, in contrast to other learned models, a learned variation operator therefore needs to be studied in combination with search dynamics, such as selection.

However, proposing non-linear structures during search does not automatically lead to high-quality solutions. Across all problems, standard GP proposes the most strongly non-linear offspring, although many of these solutions are discarded by selection. The final best programs found with GP are as non-linear as those of TSGP but perform considerably worse. What matters is therefore not how much non-linearity an operator proposes, but whether these structures are relevant for the search. The operator of TSGP-2 proposes non-linear structures that are more frequently selected during search, indicating that the variations are more meaningful, and introduces them earlier than TSGP.

Our findings highlight the importance of training data for a learned variation operator. Since a learned model reproduces the biases of its training data, a learned operator that is meant to be applied to a wide range of problems requires training data that comprehensively covers the relevant problem properties. As linearity is only one problem property, future work should investigate further kinds of biases in training data. This concerns for example other program behavior (e.g., quadratic or sinusoidal shapes) or other properties a learned operator can be biased in.
\begin{acks}
We used Claude Opus 5.5 as coding and writing assistant.
\end{acks}

\bibliographystyle{ACM-Reference-Format}
\bibliography{sources}

\newpage
\appendix
\section{Results per Problem}
\label{app:per-problem}
\begin{figure}[h]
\centering
\includegraphics[width=\linewidth]{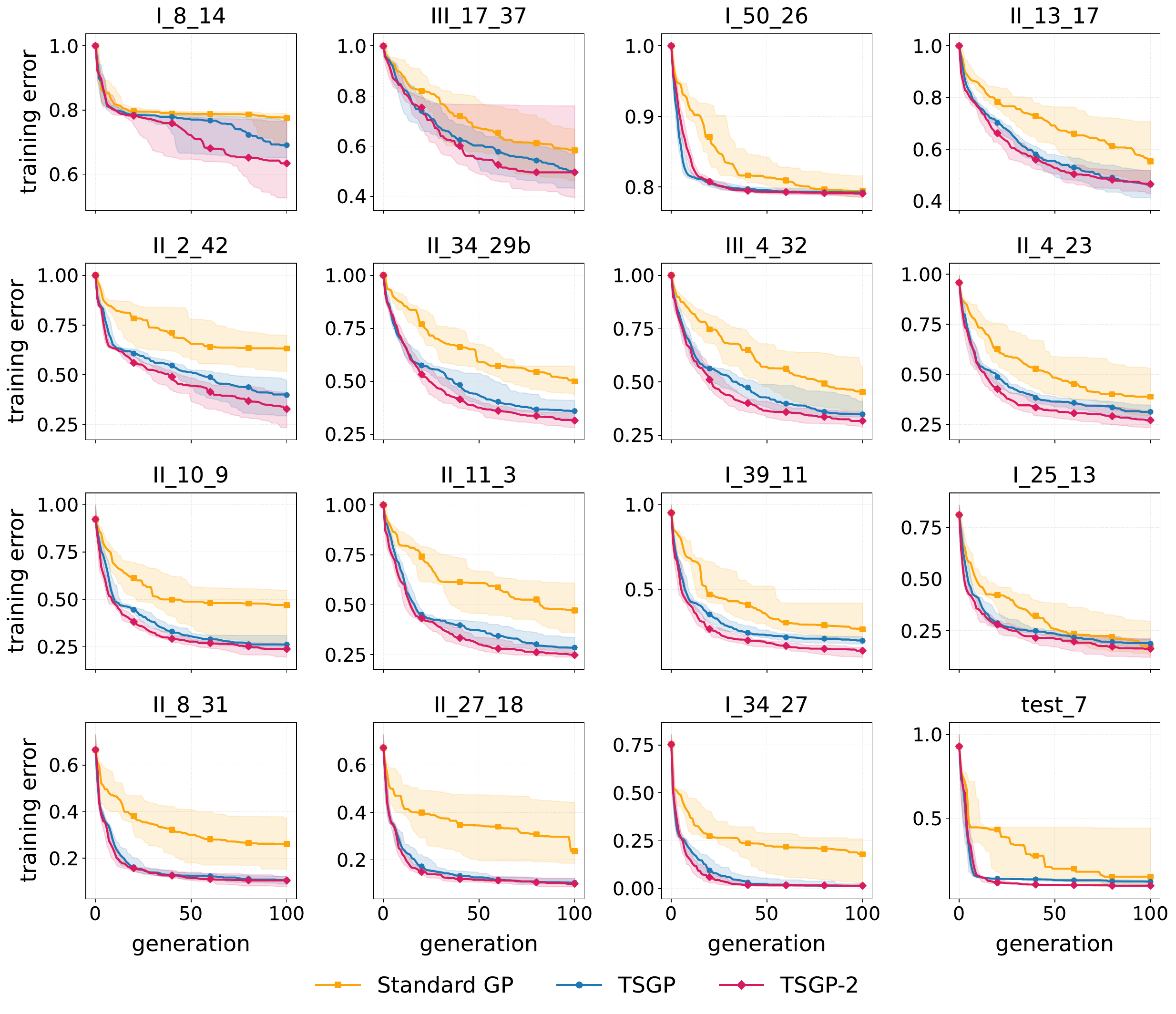}
\caption{Training error of the best program over the run on each Feynman problem. Lines are medians over the $30$ runs, bands the interquartile range.}
\label{fig:error-per-problem-feynman}
\end{figure}

\begin{figure}[p]
\centering
\includegraphics[width=\linewidth]{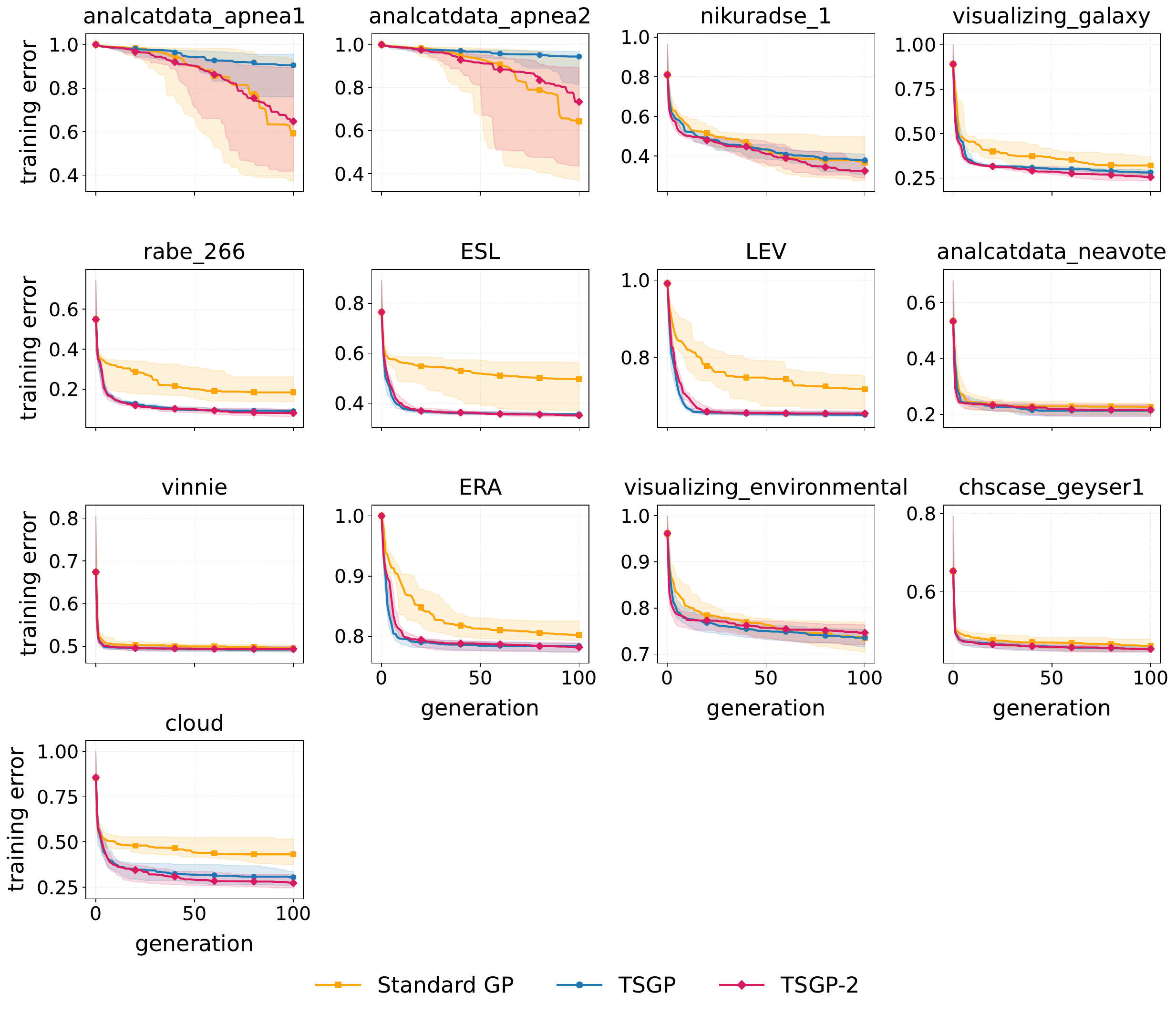}
\caption{Training error of the best program over the run on each real-world problem. Lines are medians over the $30$ runs, bands the interquartile range.}
\label{fig:error-per-problem-realworld}
\end{figure}

\begin{figure}[p]
\centering
\includegraphics[width=\linewidth]{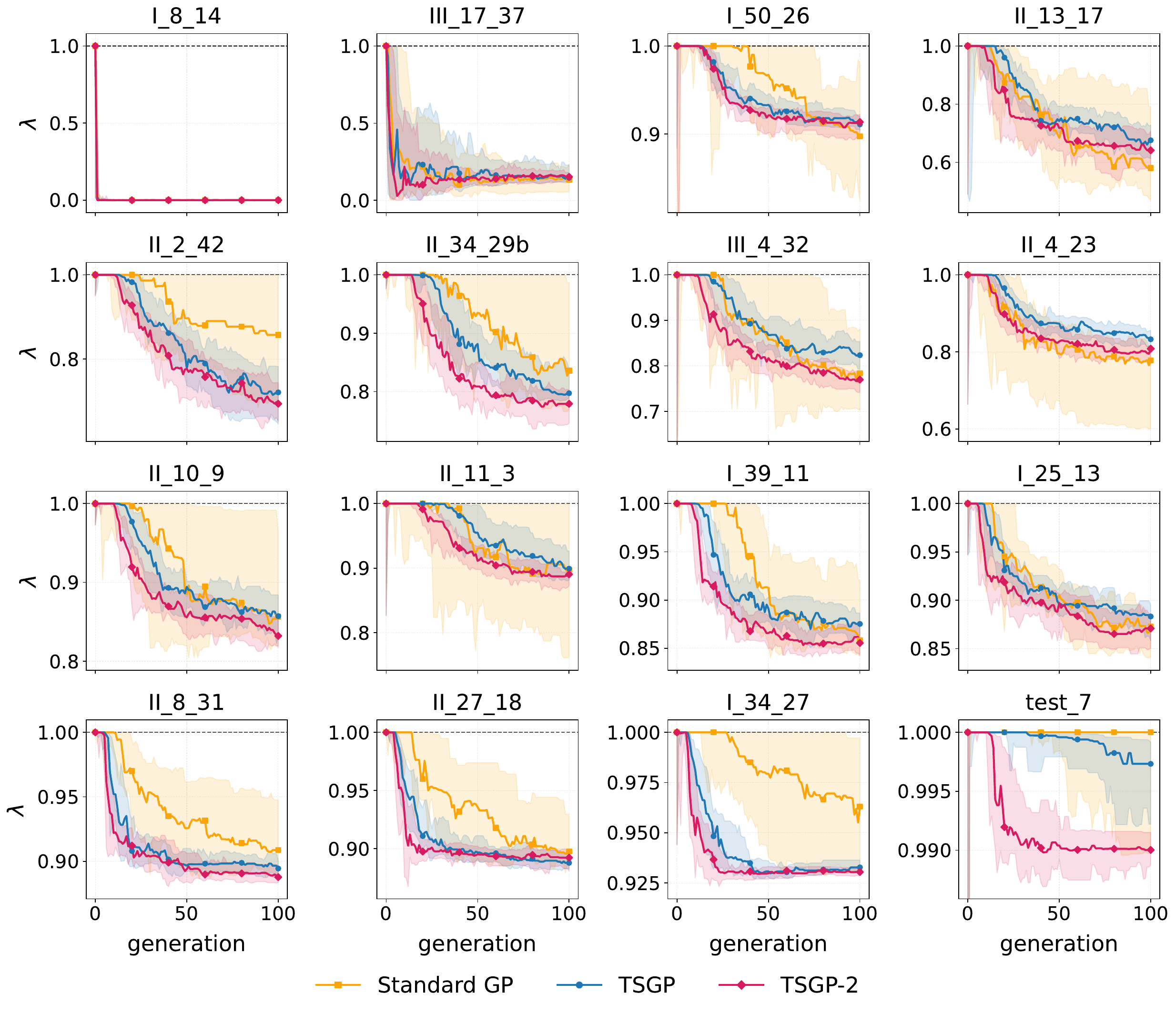}
\caption{Linearity $\lambda$ of the best program over the run on each Feynman problem. Lines are medians over the $30$ runs, bands the interquartile range.}
\label{fig:linearity-per-problem-feynman}
\end{figure}

\begin{figure}[p]
\centering
\includegraphics[width=\linewidth]{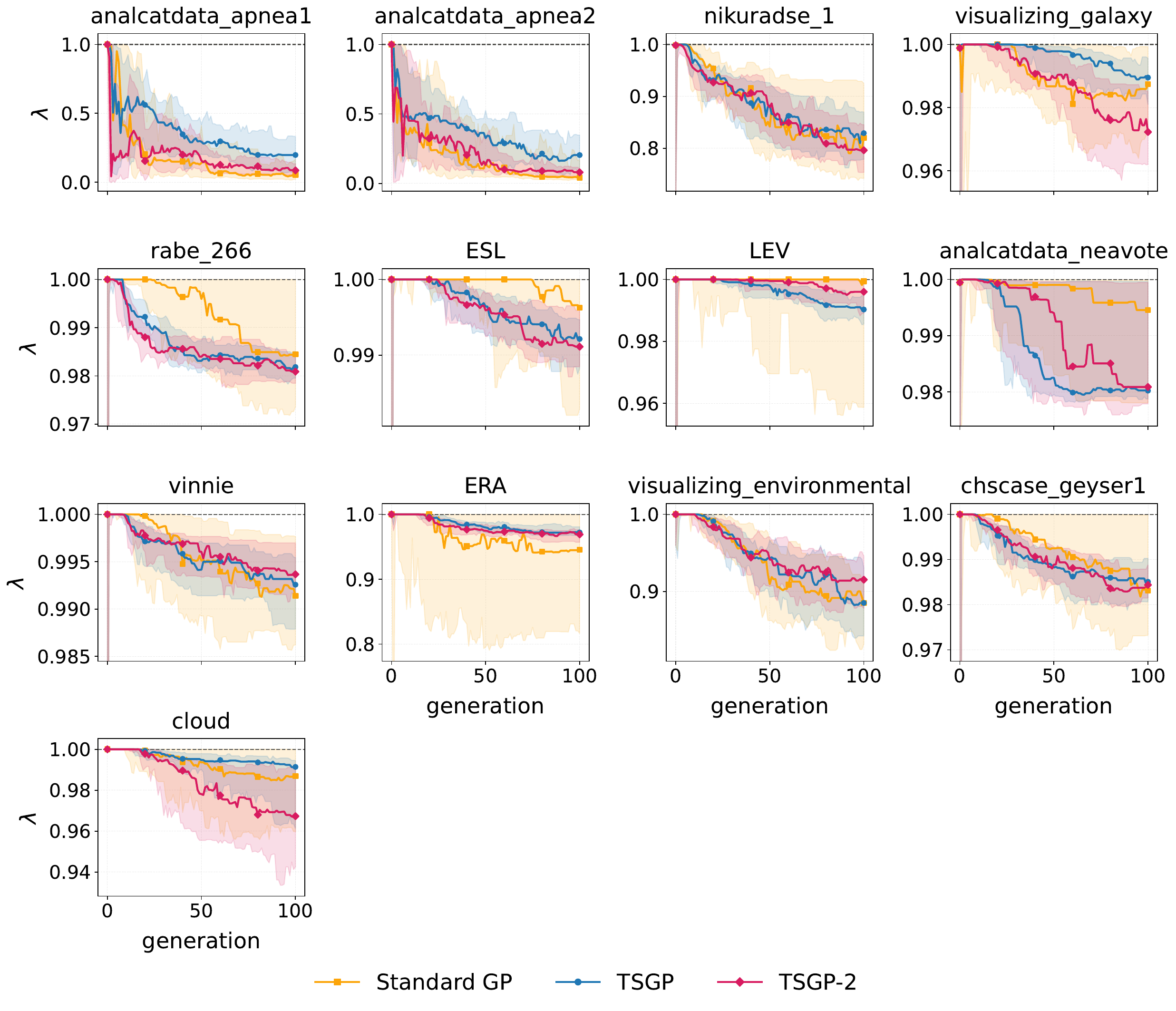}
\caption{Linearity $\lambda$ of the best program over the run on each real-world problem. Lines are medians over the $30$ runs, bands the interquartile range.}
\label{fig:linearity-per-problem-realworld}
\end{figure}

Figs.~\ref{fig:error-per-problem-feynman} to~\ref{fig:linearity-per-problem-realworld} show the training error and the linearity $\lambda$ of the best program over the run for each problem. The problems are ordered by their exploitable non-linearity (Table~\ref{tab:datasets}). Lines are medians over the $30$ runs of a problem and bands the interquartile range over these runs. 

\end{document}